\documentclass[12pt,journal,compsoc,onecolumn]{IEEEtran}
\usepackage{amsmath,epsfig}
\usepackage{lingmacros}
\usepackage{color}
 \usepackage{tree-dvips}
\usepackage{amsmath} 
\usepackage{amssymb}
\usepackage{pifont}
\usepackage[noadjust]{cite}
\usepackage{multirow}

\newlength{\bibitemsep}
\newlength{\bibparskip}
\let\oldthebibliography\thebibliography
\renewcommand\thebibliography[1]{
  \oldthebibliography{#1}
  \setlength{\parskip}{\bibitemsep}
  \setlength{\itemsep}{\bibparskip}
}

\def\x{{\mathbf x}}

\def\X{{\cal X}}
\def\XX{{\bf X}}

\def\Y{{\cal Y}}
\def\S{{\cal I}}

\def\x{{\bf x}}
\def\y{{\bf y}}
\def\D{{\cal D}}

\def\1{{\bf 1}}
\def\I{{\cal I}} 
\def\p{{p}}
\def\q{{q}}
\def\x{{\bf x}}  
\def\Y{{\cal  Y}}  

\def\Z{{\bf Z}}

\usepackage[ruled,vlined]{algorithm2e}

\title{Image augmentation with invertible networks in interactive satellite image change detection}
\author{Hichem Sahbi \\
$ $ \\
Sorbonne University, CNRS, LIP6,  F-75005, Paris, France 
 }

\begin{document}
 \maketitle
\begin{abstract}
This paper devises a novel interactive satellite image change detection algorithm based on active learning. Our framework employs an iterative process that leverages a question-and-answer model. This model queries the oracle (user) about the labels of a small subset of images (dubbed as display), and based on the oracle's responses, change detection model is dynamically updated. The main contribution of our framework resides in a novel invertible network that allows augmenting displays, by mapping them from highly nonlinear input spaces to latent ones, where augmentation transformations become linear and more tractable. The resulting augmented data are afterwards mapped back to the input space, and used to retrain more effective change detection criteria in the subsequent iterations of active learning. Experimental results demonstrate superior performance of our proposed method compared to the related work.
\end{abstract}

\section{Introduction}
\label{sec:intro}
Satellite image change detection seeks to map relevant changes that may occur on a given area over time. This task has multiple applications ranging from urban planning, to environmental monitoring, through disaster management~\cite{ref4,ref5,ref5v1,refffabc4}. Change detection is challenging due to several irrelevant variations including acquisition conditions (seasonal, atmospheric and radiometric  variations, occlusions) and sensor heterogeneity (resolution, registration and artifacts). Early change detection algorithms are based on comparisons of multi-temporal images, using differences and thresholding, vegetation indices, change vector and principal component analysis \cite{ref7,ref9,ref11,ref13,ref13v1,ref13v2}. Other existing works require a preliminary preprocessing step that mitigates the impact of irrelevant changes (geometric and radiometric calibration, noise reduction, cloud masking, etc.) \cite{ref14,ref15,ref17,ref20,refffabc8}, or handle these irrelevant changes as a part of appearance modeling using machine learning~\cite{ref21v1,ref21,ref25,ref26,refffabc7,refrefrefICIP2014,ref28,ref27,refffabc3,refffabc0}.\\

\indent While machine learning methods show significant promise in satellite image change detection, their accuracy is heavily reliant on the availability of extensive, accurately labeled training data~\cite{refffabc1,sahbi2021b}. This reliance presents a significant challenge, as comprehensive labeling of relevant and irrelevant changes is often impractical. Moreover, even with abundant hand-labeled data, labels may not perfectly align with the specific user's interpretation of changes \cite{refffabc5,refffabc6}. To address these limitations, existing works are exploring approaches that minimize the need for extensive labeled data. Augmentation~\cite{ref27vvv1,ref27vvv2,ref27vvv3,ref27vvv4,ref27vvv5,ref27vvv6,ref27vvv7,ref27vvv8,ref27vvv9}, few-shot \cite{reff45v1,reff45}  and self-supervised learning \cite{refff2v1,refff2} are promising directions, although they may not fully capture the user's specific intent. Alternatively, active learning offers a more user-centered approach~\cite{reff1,reff2,reff13,reff16,reff15,reff53,reff12,reff74,reff58,refsahbicvpr2008} where the user iteratively provides annotations for a small set of carefully selected examples, guiding the learning process towards the user's specific definition of relevant changes. This iterative process allows for the development of user-aware change detection models with minimal hand-labeling effort.\\

\indent In this paper, we introduce a novel question-and-answer (Q\&A) framework for satellite image change detection. This approach queries an oracle (e.g., domain expert) about the relevance of observed changes in images, and according to the oracle's feedback, the model dynamically adjusts its change detection criteria. While traditional supervised change detection methods rely on large, diverse labeled datasets, which can be expensive and time-consuming to acquire, our Q\&A framework minimizes the need for extensive manual labeling by allowing the oracle to provide feedback in a more efficient and targeted manner. However, this targeted labeling often results in small and potentially biased training datasets, limiting the model's generalization ability. To address this challenge, we propose a novel data augmentation technique, based on {\it invertible networks}, specifically designed to enhance the diversity and robustness of the limited training data within this Q\&A framework. The proposed method operates by first projecting training data from the original input (ambient) space into a latent space. Within this latent space, the data undergoes a disruption process. Finally, the disrupted data is projected back into the original ambient space. \textcolor{black}{Unlike existing augmentation techniques which rely on mixup or interpolation either on input spaces or low dimensional bottlenecks~(e.g., \cite{ref27vvv2,ref27vvv7}), the particularity of our method resides in its ability to explore and augment data lying on highly nonlinear manifolds, by mapping them to a latent space, where augmentations can be performed with stable and exactly invertible (bijective) networks, and also in a tractable way.} Experiments conducted on satellite image change detection, show the relevance of our proposed ``invertible network-based augmentation'' approach.

\section{Proposed Method}

Let $\I_r = \{\p_1, \dots , \p_n\}$, $\I_t = \{\q_1, \dots , \q_n\}$  denote two registered satellite images taken at two different time-stamps $t_0$ and $t_1$, with $\p_i$, $\q_i \in \mathbb{R}^d$ being an aligned patch-pair. Considering  $\I=\{\x_1,\dots, \x_n\}$, with each $\x_i=(\p_i, \q_i)$, and $\Y = \{\y_1, \dots, \y_n\}$ the unknown labels of $\I$; we seek to train a classifier $f: {\cal I} \rightarrow \{-1,+1\}$ that infers the unknown labels in $\cal Y$ with $\y_i = +1$ if the patch $\q_i$  {\it shows} a ``change'' compared to  $\p_i$, and $\y_i = -1$ otherwise. Training an accurate $f$ requires enough labeled data which are usually scarce and their hand-labeling expensive; we consider instead a {\it frugal training} regime that uses {\it as few labeled data as possible} while maintaining a high accuracy of $f$.   
\subsection{Frugal satellite image change detection}\label{frugal}
Our goal is to a achieve frugal interactive change detection using a question \& answer strategy that probes an oracle (user) about the labels of a few set of samples (dubbed as display), and according to the oracle's responses, iteratively update a change detection criterion $f$. Let $\D_t\subset \I$ denote a subset of samples shown to the oracle, at iteration $t$, whose labels $\Y_t$ are unknown; $\D_t$ is selected to make the {\it most positive impact} on $f$  when the latter is retrained with $\cup_{k=0}^t \D_k$. Starting from a random display  $\D_0$, we design our change detection criteria $f_0,\dots,f_T$ iteratively according to the subsequent steps \\

\noindent 1/ Probe the oracle about the labels of $\D_t$ and retrain a change detection criterion $f_t (.)$ on $\cup_{k=0}^t (\D_k,\Y_k)$; as shown in section~\ref{inver} (and later in experiments),  $\{f_t (.)\}_t$ correspond to invertible networks, and precisely invertible graph convnets. \\
2/ Pick the next display $\D_{t+1}\subset \S\backslash\cup_{k=0}^t \D_k$ to re-probe the oracle; it's worth emphasizing that probing the oracle with all the possible displays  $\D \subset \S\backslash\cup_{k=0}^t \D_k$, learning the underlying criteria $f_{t+1} (.)$ on $\D \cup_{k=0}^t \D_t$, evaluating their accuracy, prior to select the most accurate display, is clearly intractable. In frugal interactive learning, display selection strategies, in relation to active learning, are usually more efficient and also effective. Nonetheless, many of these heuristics are equivalent to basic display strategies that -- uniformly -- sample random displays (see for example \cite{reff2} and references within). \textcolor{black}{Hence, we consider, in this work, both random sample selection strategy and active learning one~\cite{refff33333}}, and as a contribution, we improve sample diversity using a novel augmentation method. The latter relies on invertible networks that allow exploring and enriching data (lying on nonlinear manifolds, a.k.a., ambient spaces) by (i) mapping them from ambient to latent spaces, and (ii) augmenting them using simple linear transformations in the latent space, prior to (iii) remap the transformed samples from latent to ambient spaces. This framework not only makes exploring highly nonlinear manifolds (more) tractable but also enhances the accuracy of the retrained change detection criteria. All the details of our proposed method are shown in the next two subsections which constitute the main contribution of this work.

\def\VV{{\cal V}}
\def\E{{\cal E}}
\def\A{{\bf A}}
\def\UU{{\bf U}} 
\def\W{{\bf W}}
\def\N{{\cal N}}
\def\SS{{\cal S}}
\def\FF{{\cal F}}
\subsection{Invertible networks}\label{inver}

Let $\XX$ denote a random variable standing for all possible patch-pairs taken from an existing but unknown probability distribution $P_\XX$ in an ambient space $\X \subseteq \mathbb{R}^d$. Considering $\Z$ as a latent representation associated to $\XX$ drawn from a known distribution $P_\Z$ in a latent space ${\cal Z}  \subseteq \mathbb{R}^d$; our goal is to learn an invertible diffeomorphism network $f$ from $\X$ to $\cal Z$. With this invertible $f$, data lying on a non-euclidean space $\X$ (i.e., manifold) can be mapped to an euclidean space $\cal Z$ where augmentations may easily be achieved using linear transformations, and mapped back to the ambient space. In order to capture the probability distribution $P_\XX$, we consider a reparametrization in the latent space as $f(\XX) \in \cal Z$ where $f$ models the distribution  $P_\Z$ in $\cal Z$. As described subsequently, effective augmentations can be achieved in the latent space, instead of the ambient one where nonlinear transformations are intractable.\\

\noindent Let's subsume $f$ as a multi-layered neutral network $f_{t,\theta}$, being $t$ the active learning cycle, $\theta=\{\W_1,\dots,\W_L\}$, $L$ its depth, $\W_\ell \in \mathbb{R}^{d_{\ell-1} \times d_\ell}$ its $\ell^{th}$-layer weight tensor, and $d_\ell$ the dimension of the $\ell^{th}$-layer output. In what follows, we omit both $\theta$ and $t$ in the definition of $f_{t,\theta}$ and we rewrite  a given layer $\ell$ of  $f$  as $\phi^\ell=g_\ell (\W_\ell^\top \phi^{\ell-1})$, $\ell \in \{2,\dots,L\}$, being $g_\ell$ a nonlinear activation function; without a loss of generality, we also omit the bias in the definition of $\phi^\ell$. Provided that the matrices in $\theta$ are invertible, and the activation functions $\{g_\ell\}_\ell$ bijective (such as leaky-ReLU), one may obtain a bijective and thereby an invertible network $f$ with equidimensional layers\footnote{Excepting the softmax layer whose dimensionality depends on the number of classes. Nevertheless, a simple trick guarantees constant dimensionality $d$ by adding fictitious softmax outputs to match any targeted $d$.}. This bijection property is valuable, as it reduces the complexity of achieving augmentation in the latent space instead of the ambient one. On another hand, by the invertibility of $f$, this guarantees that augmented samples, once inverted, belong to the nonlinear distribution  $P_\XX$, i.e., the manifold  enclosing data in the ambient space. \\

\def\III{{\bf I}}
\noindent {\bf Stability \& Invertibility.} The success of the generative properties of the inverted network $f^{-1}$ is reliant on the stability of $f$. Differently put, when $f$ is $M$-Lipschitzian (with $M \approx 1$), the network  $f^{-1}$ will also be $M$-Lipschitzian (with $M\approx 1$) \cite{refffabc888},  so any slight linear transformation (augmentation) of training samples in the latent space will result into a slight nonlinear transformation of these samples in the ambient space when applying $f^{-1}$. This ultimately leads to stable sample transformations in the ambient space, i.e., transformed samples follow the actual distribution  $P_\XX$ in $\cal X$. As the Lipschitz constant of $f$ is $M=\prod_{\ell} \|\W_\ell \|_2 . \big|g'_\ell\big|$, the sufficient conditions guaranteeing that both $f$ and $f^{-1}$ are $M$-Lipschitzian (with $M\approx 1$) corresponds to (i) $\|\W_\ell \|_2 \approx 1$, and (ii) $\big|g'_\ell\big| \approx 1$ for all $\ell$. Hence, by design, conditions (i)+(ii) could be satisfied by choosing the slope of the activation functions to be close to one (in practice to 0.99 and 0.95 respectively for the positive and negative orthants of leaky-ReLU), and also by constraining all the weight matrices in $\theta$ to be {\it orthonormal} which also guarantees their invertibility. This is obtained by considering the following loss when training $f$
\begin{equation}\label{ce}
\min_{\{\W_\ell\}_\ell}{\textrm{CE}}(f;\{\W_\ell\}_\ell) + \lambda \ \sum_{\ell} \big\|\W_\ell^\top \W_\ell-\III\big\|_F,
\end{equation}
here CE stands for cross entropy, $\III$ for identity, $\|.\|_F$ for the Frobenius norm and $\lambda>0$ (with $\lambda=\frac{1}{d}$ in practice); if $\W_\ell^\top \W_\ell-\III =0$, then  $\W_\ell^{-1}=\W_\ell^\top$ and  $\|\W_\ell \|_2 =\|\W_\ell^{-1} \|_2=1$.  When optimizing Eq.~\ref{ce}, the learned networks are guaranteed to be discriminative, stable and invertible. With the loss in Eq.~\ref{ce}, the gain in the discriminative and the generative performances of the learned network $f$ is noticeable (particularly in the frugal data regime) as shown later through experiments in interactive satellite image change detection. 

\subsection{Augmentation}
Training invertible networks may result into overfitting in the frugal data regimes. To circumvent this issue, we augment our displays $\cup_{k=0}^t \D_k$ for each active learning cycle $t$. Our conjecture considers that disrupting data in $\cup_{k=0}^t \D_k$ (either individually or pairwise as shown subsequently) allows generating new samples that inherit the same labels.  Considering the union of all displays $\cup_{k=0}^t \D_k$; we train an invertible network $f$, prior to {\it invert} transformed displays in the ambient space using $f^{-1}$. Note that these transformations  while being linear in the latent space, are highly nonlinear in the ambient space, due to the nonlinearity of $f^{-1}$, and thereby capture better the distribution $P_\XX$. Hence, one may explore and augment samples in nonlinear manifolds in a much easier way thanks to the invertibility of $f$. \\

\indent Considering the aforementioned goal, we introduce two different augmentation methods based on unary and binary transformations.  Unary augmentations disrupt training samples by adding random noises with increasing amplitudes in the latent space, while preserving the underlying labels. Binary augmentations are obtained as linear interpolations in the latent space, which are applied only on sample pairs with the same labels resulting into augmented samples that inherit the same labels. In what follows, we further detail these transformations that allow retraining more effective invertible networks, and thereby improving the performances of interactive change detection.\\

\noindent {\bf Unary Augmentations.}  Given a sample $\x \in \cup_{k=0}^t \D_k$ whose known label is $\y$, the pair $(\x,\y)$ is disrupted by adding a multivariate normal noise $v \sim {\cal N}(0_d,{\bf I})$ as
\begin{equation}
\hat{\x} = f^{-1}(f(\x)+ \delta . v),
\end{equation} 
here $0_d$ stands for zero mean and $\delta >0$. With this perturbation, and due to the stability and invertibility of $f$, the generated sample $\hat{\x}$ inherits the same label $\y$ as the original $\x$.\\

\noindent {\bf Binary Augmentations.} Considering two samples $\x_1$ and $\x_2$ in $ \in \cup_{k=0}^t \D_k$ with the same known labels, we generate a new sample $\hat{\x}$ as a convex combination of the latent representations of $\x_1$ and $\x_2$. The new sample $\hat{\x}$ inherits the same label as $\x_1$ and $\x_2$. 
\begin{equation}
\hat{\x} = f^{-1}\bigg[\big(v_1 \odot f(\x_1)+  v_2 \odot f(\x_2)\big) \oslash (v_1+v_2)\bigg],
\end{equation} 
being  $v_1 \sim {\cal N}(0_d,{\bf I})$, $v_2 \sim {\cal N}(0_d,{\bf I})$, and the entrywise multiplication and division (resp. denoted as $\odot$  and $\oslash$) guarantee that each dimension in the latent space corresponds to a convex combination. Two variants of this binary augmentation are considered; in the first one, the dimensions of the random perturbation vectors $v_1\oslash (v_1+v_2)$ and $v_2\oslash (v_1+v_2) \in [0,1]^d$ are thresholded to obtain quantized values in $\{0,1\}^d$ in the convex combination\footnote{Each value in each dimension is quantized to 1 if its initial value is larger than 0.5 and to 0 otherwise.} whilst in the second variant, all the dimensions are kept unchanged. The latter variant generates {\it soft} combinations of dimensions in $f(\x_1)$, $f(\x_2)$ whereas the former implements {\it crisp} combinations. As shown in experiments, the crisp variant is more effective (compared to the soft one) and provides enough variability in the augmented training samples.

\section{Experiments}

Change detection experiments were conducted on the Jefferson dataset, comprising 2,200 non-overlapping patch pairs. Each patch pair consists of two 30x30 RGB pixels extracted from registered (bi-temporal) GeoEye-1 satellite images of Jefferson, Alabama. These images, acquired in 2010 and 2011 with a spatial resolution of 1.65 meters per pixel, exhibit various changes, including those caused by tornadoes (e.g., building destruction), as well as numerous no-change instances (including irrelevant ones such as cloud cover). The ground truth labels reveal a severe class imbalance, with 2,161 negative (no/irrelevant change) pairs and only 39 positive (relevant change) pairs, indicating that over 98\% of the area exhibits no change. This high class imbalance poses a significant challenge for accurately identifying relevant changes. In our experiments, half of the dataset was used for training, while the remaining half was reserved for evaluation. Due to the highly imbalanced class distribution, we show model performance using the Equal Error Rate (EER) on the evaluation set. Lower EER values indicate better performance. \\
\noindent In order to classify these patch pairs as change / no-change, we use a graph convnet as backbone to train our invertible network. The architecture of our graph convnet comprises eight ``mono-head attentions + (32 filters) convolutions'' layers followed by one fully connected and a softmax layer. Again, we train this network using the loss in Eq.~\ref{ce} and the setting discussed in section~\ref{inver} in order to make it invertible and stable.    

\subsection{Model Analysis \& Ablation}

Table~\ref{tab12} shows a  comparison of change detection EERs w.r.t. different ``unary'' and ``binary'' augmentation settings. We observe that the best average (AUC) performances are obtained by ``unary'' with $\delta=1.00$  corresponding to the largest data perturbation which improves the diversity of the displays. Table~\ref{abl} shows an ablation study of different criteria used in our method, namely ``display'' (``random'' vs. ``optimized''~\cite{refff33333}), and ``space'' (``ambient'' vs. ``latent''\footnote{i.e., data perturbations are achieved either (i) directly in the ambient space or (ii) first in the latent space and afterwards inverted.}).  From these results, we observe the highest impact of latent space augmentation particularly when the display model is optimized. Nonetheless, the impact of the latent space is more pronounced compared to the impact of display optimization; in other words, the gain when using the latent space  is important both with optimized and random displays. All  these EER performances are shown  for different sampling percentages defined --- at each iteration $t$ --- as $(\sum_{k}^t |\D_k|/(|\I|/2))\times 100$  with again $|\I|=2,200$ and $|\D_k|$ set to $16$.

 \begin{table}\label{ma}
 \resizebox{1.01\columnwidth}{!}{
 \begin{tabular}{c||ccccccccc||c}
 Method / Iter & 2 & 3& 4& 5& 6& 7& 8& 9 & 10 & AUC. \\
 \hline
Unary  ($\delta=0.01$) &   17.72 &   8.18  &  4.00   &   3.63  &    3.27 &    3.00 &    2.18 &    2.09  &   1.45   & 5.06 \\
Unary  ($\delta=0.10$)  &   22.18 &  13.81  &  8.54   &   4.90  &    2.72 &   3.00  &    2.63 &    2.54  &    2.27  & 6.95  \\ 
Unary  ($\delta=1.00$)  &  21.00  &   6.27  &  3.18   &    2.90 &    2.18 &    1.54 &    1.63 &    1.54  &    1.72  & \bf4.66 \\
Binary  (Soft)           &   16.00 &   15.18 &  4.63 &    3.63 &  3.90   &    3.81 &    3.27 &    3.63  &    3.45 &  6.39 \\
Binary  (Crisp)        &  16.63    &    7.54 &  4.18 &    4.00 &    3.18 &    2.09 &    2.09 &    2.09  &    1.90 & 4.85 \\
   \hline
Samp\%   &2.90 & 4.36& 5.81& 7.27& 8.72& 10.18& 11.63& 13.09 & 14.54 & - 
\end{tabular}}
 \caption{\textcolor{black}{This table shows a comparison of change detection EERs w.r.t. unary and binary augmentation methods with different settings. These results are shown for different iterations $t$ (Iter) and the underlying sampling rates (Samp). The AUC (Area Under Curve) corresponds to the average of EERs across iterations.}}\label{tab12} 
\end{table}

 \begin{table}
 \resizebox{1.01\columnwidth}{!}{
 \begin{tabular}{cc||ccccccccc||c}
 Display & Space & 2 & 3& 4& 5& 6& 7& 8& 9 & 10 & AUC. \\
 \hline
 random  & ambient & 16.27   &   12.36 &    6.00 &    4.18 &    5.54 &    4.81 &    4.45 &    4.00  &    3.00  & 6.73 \\
optimized  & ambient  &  20.45  &   9.72  &  4.54   &    2.90 &    2.63 &    2.90 &    2.81 &    2.81  &    2.81  & 5.73 \\     
random  & latent  &  11.90  &   7.72  &  3.81   &    3.00 &    4.45 &    4.36 &    3.90 &    3.90  &    3.27  & 5.15 \\
optimized  & latent  &  21.00  &   6.27  &  3.18   &    2.90 &    2.18 &    1.54 &    1.63 &    1.54  &    1.72  & \bf4.66 
\end{tabular}}
 \caption{\textcolor{black}{This table shows an ablation study of our augmentation method. These results correspond to the best setting in table~\ref{tab12}, i.e., with unary augmentation and $\delta=1.00$.}}\label{abl} 
 \end{table} 

 \subsection{Comparison}\label{compare}
 
\begin{figure}[tbp]
  \centering 
\includegraphics[width=0.70\linewidth]{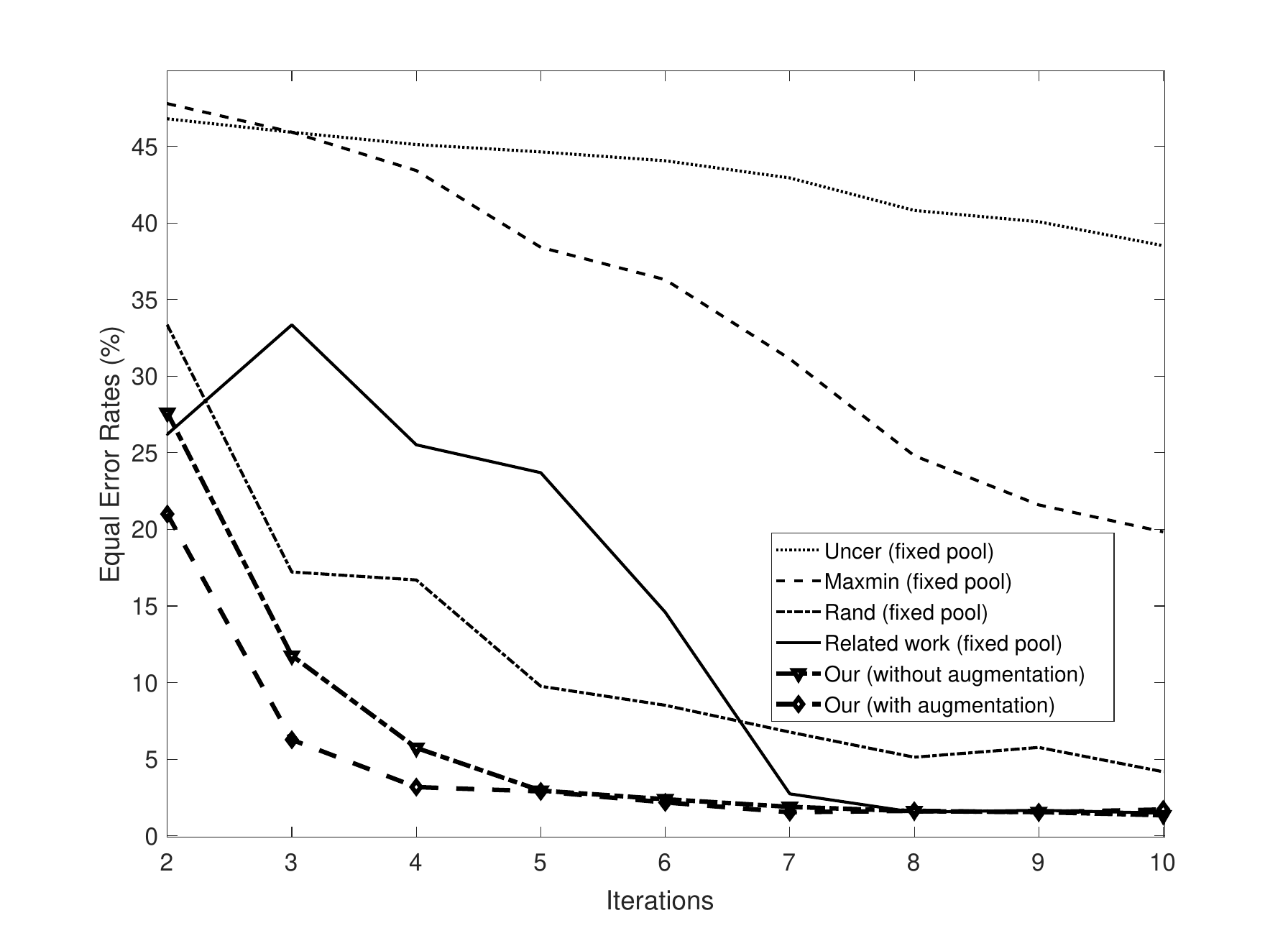}
 \caption{This figure shows a comparison of different sample selection strategies w.r.t. different iterations (Iter) and the underlying sampling rates in table~\ref{tab12} (Samp). Here Uncer and Rand stand for uncertainty and random sample  selection respectively. Note that fully-supervised learning achieves an EER of $0.94 \%$.  Related work stands for the method in \cite{refff33333}; see again section~\ref{compare} for more details. ``Our (with augmentation)'' stands for the best setting in table~\ref{tab12} and ``Our (without augmentation)'' stands for the same setting but without augmentation.}\label{tab2}\end{figure}

 In order to further evaluate the effectiveness of our proposed approach in change detection,  we compare its performance against other active learning sample selections. These include a totally {\it random} search strategy, which selects samples randomly from the pool of unlabeled data, and two optimized strategies (i) {\it maxmin} which  aims to maximize the minimum distance between selected samples, ensuring diversity in the chosen data, and (ii) {\it uncertainty} which prioritizes samples for which the classifier exhibits the highest uncertainty, typically measured by the proximity of the classifier's predicted scores to its class boundary values. We further compare our method against the highly optimized strategy in \cite{refff33333}. This strategy assigns a probability measure to the entire unlabeled dataset and then selects the display with the highest probability. Additionally, we established an upper bound on performance by training our invertible network on the complete labeled dataset, using the ground-truth annotations for {\it all} samples. The EER performance, plotted (in Figure~\ref{tab2}) against different iterations and sampling rates, demonstrates the superior performance of our proposed method compared to the previously discussed sample selection strategies. \textcolor{black}{With the exception of \cite{refff33333}, most of these comparative methods exhibit limited ability to effectively identify the rare change class. Whereas {\it random and maxmin} effectively capture data diversity in the initial stages of the interactive search process, they fail to refine the decision function effectively in later iterations. In contrast, {\it uncertainty} effectively refines the decision function, but often lacks diversity in the selected samples. The display strategy in \cite{refff33333} effectively combines the advantages of {\it random, maxmin, and uncertainty} sample selections. However, this approach suffers from a key limitation: the selection of displays is restricted to a limited set of training data. In contrast, our method dynamically expands the training data (displays) by exploring and augmenting them within the manifold that encapsulates their distribution. This data augmentation proves particularly beneficial in interactive change detection particularly in highly frugal labeling regimes.}

\section{Conclusion}
This paper introduces a novel interactive change detection algorithm based on active learning. A key strength of our approach lies in the diversity of our displays, which are augmented using a novel invertible network. This augmentation process is achieved by mapping --- via our invertible network --- training data from highly nonlinear input spaces to latent ones where augmentations become linear and much more tractable. These augmented samples are mapped back to the input space thanks to the invertibility of our network, and used to iteratively retrain more effective change detection criteria. Experiments show  superior performance of our model compared to the related work.


\newpage 
  

{   
\footnotesize 
\begin{thebibliography}{1}
  \bibitem{refffabc1}  P. Vo and H. Sahbi. "Transductive kernel map learning and its application to image annotation." BMVC. 2012.

\bibitem{ref4}  D. Brunner, G. Lemoine, and L. Bruzzone, Earthquake damage assessment of buildings using vhr optical and sar imagery, IEEE Trans. Geosc. Remote Sens., vol. 48, no. 5, pp. 2403--2420, 2010.

\bibitem{ref5} H. Gokon, J. Post, E. Stein, S. Martinis, A. Twele, M. Muck, C. Geiss, S. Koshimura, and M. Matsuoka, A method for detecting buildings destroyed by the 2011 tohoku earthquake and tsunami using multitemporal terrasar-x data, GRSL, vol. 12, no. 6, pp. 1277--1281, 2015.

\bibitem{ref5v1}  Wen, Y. et al. GCD-DDPM: A generative change detection model based on difference-feature guided DDPM. IEEE Transactions on Geoscience and Remote Sensing, 2024

\bibitem{refffabc0} Q. Oliveau and H. Sahbi. "Learning attribute representations for remote sensing ship category classification." IEEE JSTARS 10.6 (2017): 2830-2840.


\bibitem{ref7} J. Deng, K. Wang, Y. Deng, and G. Qi, PCA-based land-use change detection and analysis using multitemporal and multisensor satellite data, IJRS, vol. 29, no. 16, pp. 4823--4838, 2008.
 \bibitem{ref9} R. Radke, S. Andra, O. Al-Kofahi, and B. Roysam, Image change detection algorithms: A systematic survey, IEEE Trans. on Im Proc, vol. 14, no. 3, pp. 294--307, 2005.
 
  \bibitem{aaaaa2}  M.  Jiu and  H.  Sahbi. Nonlinear deep kernel learning for image annotation.  IEEE Transactions on Image Processing 26 (4), 1820-1832.
 
 
 \bibitem{ref11} S. Liu, L. Bruzzone, F. Bovolo, M. Zanetti, and P. Du, Sequential spectral change vector analysis for iteratively discovering and detecting multiple changes in hyperspectral images, TGRS,  53(8), pp. 4363--4378, 2015.
\bibitem{ref13} G. Chen, G. J. Hay, L. M. Carvalho, and M. A. Wulder, Object-based change detection, IJRS, vol. 33, no. 14, pp. 4434--4457, 2012.

\bibitem{ref13v1} LI, Liangliang, MA, Hongbing, ZHANG, Xueyu, et al. Synthetic aperture radar image change detection based on principal component analysis and two-level clustering. Remote Sensing, vol. 16, no 11, p. 1861, 2024.

\bibitem{ref13v2} LISTIANI, Indira Aprilia, ZANETTI, Massimo, et BOVOLO, Francesca. Time Series Directional Change Vector Analysis. In IEEE IGARSS 2024-2024, p. 8683-8686, 2024.

\bibitem{refffabc5} H. Sahbi. "Interactive satellite image change detection with context-aware canonical correlation analysis." IEEE GRSL, (14)5, 2017.
  



\bibitem{ref14} J. Zhu, Q. Guo, D. Li, and T. C. Harmon, Reducing mis-registration and shadow effects on change detection in wetlands, Photogrammetric Engineering \& Remote Sensing, vol. 77, no. 4, pp. 325--334, 2011.

\bibitem{ref15} A. Fournier, P. Weiss, L. Blanc-Fraud, and G. Aubert, A contrast equalization procedure for change detection algorithms: applications to remotely sensed images of urban areas, In ICPR, 2008

\bibitem{refffabc6}  H. Sahbi. "Relevance feedback for satellite image change detection." IEEE ICASSP, 2013.

\bibitem{ref17} Carlotto, Detecting change in images with parallax, In Society of Photo-Optical Instrumentation Engineers, 2007 

 \bibitem{aaaaa7}  F.   Yuan,  G-S.  Xia,  H. Sahbi,  V. Prinet.  Mid-level features and spatio-temporal context for activity recognition.  Pattern Recognition 45 (12), 4182-4191



\bibitem{ref20}  S. Leprince, S. Barbot, F. Ayoub, and J.-P. Avouac, Automatic and precise orthorectification, coregistration, and subpixel correlation of
satellite images, application to ground deformation measurements,TGRS, vol. 45, no. 6, pp. 1529--1558, 2007.

\bibitem{ref21v1}  Cheng, Guangliang, et al. "Change detection methods for remote sensing in the last decade: A comprehensive review." Remote Sensing 16.13: 2355., 2024.

\bibitem{ref21} Pollard, Comprehensive 3d change detection using volumetric appearance modeling, Phd, Brown University, 2009.
\bibitem{ref25}A. A. Nielsen, The regularized iteratively reweighted mad method for change detection in multi-and hyperspectral data, IEEE Transactions on Image processing, vol. 16, no. 2, pp. 463--478, 2007.
\bibitem{ref26} C. Wu, B. Du, and L. Zhang, Slow feature analysis for change detection in multispectral imagery, TGRS, vol. 52, no. 5, pp. 2858--2874, 2014
  

\bibitem{refffabc7} N. Bourdis, D. Marraud and H. Sahbi. "Constrained optical flow for aerial image change detection." in IEEE IGARSS, 2011.

 \bibitem{refrefrefICIP2014}  L. Wang, H. Sahbi. Bags-of-daglets for action recognition. In IEEE ICIP 2014.

\bibitem{ref28} J. Im, J. Jensen, and J. Tullis, Object-based change detection using correlation image analysis and image seg, IJRS, 29(2), 399--423, 2008.
  \bibitem{ref27} N. Bourdis, D. Marraud, and H. Sahbi, Spatio-temporal interaction for aerial video change detection, in IGARSS, 2012, pp. 2253--2256 


  \bibitem{ref27vvv1} S. Yun et al. “Cutmix: Regularization strategy to train strong classifiers with localizable features,” in In IEEE/CVF ICCV, 2019, pp. 6023–6032.
    
 \bibitem{ref27vvv2}    H. Zhang et al. “mixup: Beyond empirical risk minimization,”arXiv preprint arXiv:1710.09412, 2017.

   

\bibitem{ref27vvv3}  O. Adedeji et al. “Image augmentation for satellite images,” arXiv preprint arXiv:2207.14580, 2022.

\bibitem{ref27vvv4}  G. Osada et al. “Regularization with latent space virtual adversarial training,” in ECCV 2020. Springer, 2020, pp. 565–581.

\bibitem{ref27vvv5}  J. Howe, K. Pula, and A. A Reite, “Conditional generative adversarial networks for data augmentation and adaptation in remotely sensed imagery,” in Applications of Machine Learning. SPIE, 2019, vol. 11139, pp. 119–131.
  
\bibitem{ref27vvv6}  O-K. Yüksel et al. “Semantic perturbations with normalizing flows for improved generalization,” in In IEEE/CVF ICCV, 2021, pp. 6619–6629.

\bibitem{ref27vvv7}  G. Osada, B. Ahsan, and T. Nishide, “Mixed samples data augmentation with replacing latent vector components in normalizing flow,” in First Workshop on Interpolation Regularizers and Beyond at NeurIPS 2022, 2022.

\bibitem{ref27vvv8}  C. Beckham et al. “On adversarial mixup resynthesis,” Advances in neural information processing systems, vol. 32, 2019.

\bibitem{ref27vvv9}  D. Berthelot et al. “Understanding and improving interpolation in autoencoders via an adversarial regularizer,” arXiv preprint arXiv:1807.07543, 2018.


    
  \bibitem{reff45v1}   Qiu, Chunping, et al. "Few-shot remote sensing image scene classification: Recent advances, new baselines, and future trends." ISPRS Journal of Photogrammetry and Remote Sensing 209 (2024): 368-382.


\bibitem{reff45} Vinyals et al., Matching networks for one shot learning. 2016. 

 
\bibitem{reff1} Dasgupta, Sanjoy. "Analysis of a greedy active learning strategy." Advances in neural information processing systems 17 (2004).  
\bibitem{refffabc4} H. Sahbi. "Coarse-to-fine deep kernel networks." IEEE ICCV-W, 2017.

\bibitem{reff2} Burr, Settles. "Active learning." Synthesis Lectures on Artificial Intelligence and Machine Learning 6.1 (2012).
 


\bibitem{reff13}  Tianxu et al., An Active Learning Approach with Uncertainty, Representativeness, and Diversity,


 

\bibitem{reff16}  Joshi et al., Multi-class active learning for image classification. 2009. 
 

 
\bibitem{reff15} Settles \& Craven. An analysis of active learning strategies for sequence labeling tasks. 2008.

 

\bibitem{reff53} Houlsby et al., Bayesian active learning for classification and preference learning. 2011. 

 

\bibitem{reff12}  Campbell \& Broderick, Automated scalable Bayesian inference via Hilbert coresets. 2019.

 
\bibitem{reff74} Gal et al., Deep bayesian active learning with image data. 2017 
 

 
\bibitem{reff58} Pang et al., Meta-Learning Transferable Active Learning Policies by Deep Reinforcement Learning 
 
  
\bibitem{refffabc3} M. Jiu and H. Sahbi. "Laplacian deep kernel learning for image annotation." IEEE ICASSP, 2016.

\bibitem{refff2v1}   Zhao, Maofan, et al. "Beyond Pixel-Level Annotation: Exploring Self-Supervised Learning for Change Detection With Image-Level Supervision." IEEE Transactions on Geoscience and Remote Sensing (2024).

  
\bibitem{refff2}  A. Kolesnikov, X. Zhai, L. Beyer. Revisiting Self-Supervised Visual Representation Learning. IEEE/CVF CVPR, 2019, pp. 1920-1929



\bibitem{refffabc}
  Creswell, Antonia, et al. "Generative adversarial networks: An overview." IEEE Signal Processing Magazine 35.1 (2018): 53-65.

  
\bibitem{refffabc8} N. Bourdis, D. Marraud and H. Sahbi. "Camera pose estimation using visual servoing for aerial video change detection." IEEE IGARSS 2012.
 


\bibitem{Deng2009} Jia Deng, Wei Dong, Richard Socher, Li-Jia Li, Kai Li, and Li Fei-Fei.  Imagenet: A large-scale hierarchical image database. In 2009 IEEE conference on computer vision and pattern recognition, pages 248–255. Ieee, 2009

 \bibitem{refref21} H. Sahbi and N. Boujemaa. "Robust matching by dynamic space warping for accurate face recognition." Proceedings 2001 International Conference on Image Processing (Cat. No. 01CH37205). Vol. 1. IEEE, 2001.
 

\bibitem{Conf2008} Sabrina Tollari, Philippe Mulhem, Marin Ferecatu, Hervé Glotin, Marcin Detyniecki, Patrick Gallinari, H. Sahbi, and Zhong-Qiu Zhao. "A comparative study of diversity methods for hybrid text and image retrieval approaches." In Workshop of the Cross-Language Evaluation Forum for European Languages, pp. 585-592. Springer, Berlin, Heidelberg, 2008.
 
  \bibitem{aaaaa9}  H. Sahbi.  A particular Gaussian mixture model for clustering and its application to image retrieval.   Soft Computing 12 (7), 667-676

\bibitem{Ashish2017} 
Ashish Vaswani, Noam Shazeer, Niki Parmar, Jakob Uszkoreit, Llion Jones, Aidan N. Gomez, Lukasz Kaiser, Illia Polosukhin. Attention Is All You Need.  arXiv:1706.03762. 2017. 

\bibitem{Krizhevsky2012} 
Krizhevsky, Alex, Ilya Sutskever, and Geoffrey E. Hinton. "Imagenet classification with deep convolutional neural networks." Advances in neural information processing systems 25 (2012): 1097-1105.

\bibitem{refref22} A. Mazari and H. Sahbi. "MLGCN: Multi-Laplacian graph convolutional networks for human action recognition." The British Machine Vision Conference (BMVC). 2019.



\bibitem{Szegedy2016} 
Szegedy, Christian, et al. "Rethinking the inception architecture for computer vision." Proceedings of the IEEE conference on computer vision and pattern recognition. 2016.

\bibitem{theref2008} 
M.  Ferecatu and H. Sahbi.  "TELECOM ParisTech at ImageClefphoto 2008: Bi-Modal Text and Image Retrieval with Diversity Enhancement." CLEF (Working Notes). 2008.

\bibitem{Szegedy2017} 
Szegedy, Christian, et al. "Inception-v4, inception-resnet and the impact of residual connections on learning." Thirty-first AAAI conference on artificial intelligence. 2017.


  
 \bibitem{TL1}  Clemens-Alexander Brust, Christoph Kading, and Joachim Denzler.  Active learning for deep object detection. arXiv preprint arXiv:1809.09875, 2018.

   \bibitem{DA1}  Mei Wang and Weihong Deng. Deep visual domain adaptation: A survey. Neurocomputing, 312:135–153, 2018.


   \bibitem{DA2} Connor Shorten and Taghi M Khoshgoftaar. A survey on image data aug- mentation for deep learning. Journal of Big Data, 6(1):1–48, 2019.


 
   \bibitem{SS1} Keze Wang, Liang Lin, Xiaopeng Yan, Ziliang Chen, Dongyu Zhang, and Lei Zhang. Cost-effective object detection: Active sample mining with switchable selection criteria. CoRR, abs/1807.00147, 2018.

   \bibitem{SGT1} Vladimir Haltakov, Christian Unger, and Slobodan Ilic. Framework for gen- eration of synthetic ground truth data for driver assistance applications. In German conference on pattern recognition, pages 323–332. Springer, 2013.
   \bibitem{cvpr2008aaaa} 
H. Sahbi, Jean-Yves Audibert, Jaonary Rabarisoa, and Renaud Keriven. "Context-dependent kernel design for object matching and recognition." In 2008 IEEE Conference on Computer Vision and Pattern Recognition, pp. 1-8. IEEE, 2008.



   \bibitem{AL9} Begum Demir, Claudio Persello, and Lorenzo Bruzzone. Batch-mode active-learning methods for the interactive classification of remote sens- ing images. IEEE Transactions on Geoscience and Remote Sensing, 49(3):1014–1031, 2010.
 \bibitem{ICCV2007sahbi}
H. Sahbi, Jean-Yves Audibert, and Renaud Keriven. "Graph-cut transducers for relevance feedback in content based image retrieval." 2007 IEEE 11th International Conference on Computer Vision. IEEE, 2007.

 \bibitem{EXPEXP} Krause, Andreas, and Carlos Guestrin. "Nonmyopic active learning of gaussian processes: an exploration-exploitation approach." Proceedings of the 24th international conference on Machine learning. 2007.
\bibitem{refref17}  M. Jiu and H.  Sahbi. "Deep representation design from deep kernel networks." Pattern Recognition 88 (2019): 447-457.

\bibitem{refff33333} H. Sahbi, S. Deschamps, A. Stoian. Frugal Learning for Interactive Satellite Image Change Detection. IEEE IGARSS, 2021.

 \bibitem{M1} Robert Pinsler, Jonathan Gordon, Eric T. Nalisnick, and Jose Miguel Hernandez-Lobato. Bayesian batch active learning as sparse subset approximation. In Hanna M. Wallach, Hugo Larochelle, Alina Beygelzimer, Florence d’Alche Buc, Emily B. Fox, and Roman Garnett, editors, NeurIPS, pages 6356–6367, 2019.

 \bibitem{M111111} S. Thiemert,  H. Sahbi, and M. Steinebach. "Applying interest operators in semi-fragile video watermarking." Security, Steganography, and Watermarking of Multimedia Contents VII. Vol. 5681. SPIE, 2005.

 \bibitem{M2}  Ricardo BC Prudencio and Teresa B Ludermir. Selective generation of training examples in active meta-learning. International Journal of Hybrid Intelligent Systems, 5(2):59–70, 2008.

 \bibitem{M3}  Hiranmayi Ranganathan, Hemanth Venkateswara, Shayok Chakraborty, and Sethuraman Panchanathan. Deep active learning for image classification. In 2017 IEEE International Conference on Image Processing (ICIP), pages 3934–3938. IEEE, 2017.

  \bibitem{FSL} Snell, Jake, Kevin Swersky, and Richard S. Zemel. "Prototypical networks for few-shot learning." arXiv preprint arXiv:1703.05175 (2017).
  
  \bibitem{ref2015aaaa}  H. Sahbi. "Imageclef annotation with explicit context-aware kernel maps." International Journal of Multimedia Information Retrieval 4.2 (2015): 113-128.
  
\bibitem{sener2017}  Sener, Ozan, and Silvio Savarese. "Active learning for convolutional neural networks: A core-set approach." arXiv preprint arXiv:1708.00489 (2017).
  
 \bibitem{david1994}  David D Lewis and William A Gale. A sequential algorithm for training text classifiers. In SIGIR’94, pages 3–12. Springer, 1994.
  
  \bibitem{refref26}  H. Sahbi. "Learning laplacians in chebyshev graph convolutional networks." Proceedings of the IEEE/CVF International Conference on Computer Vision.  2021.


  \bibitem{Xin2013} Xin Li and Yuhong Guo. Adaptive active learning for image classification. In CVPR, pages 859–866. IEEE Computer Society, 2013.
  
  \bibitem{Ajay2009}   Ajay J. Joshi, Fatih Porikli, and Nikolaos Papanikolopoulos. Multi-class active learning for image classification. In CVPR, pages 2372–2379. IEEE Computer Society, 2009.

     \bibitem{refref16}  L. Wang and H.  Sahbi. "Bags-of-daglets for action recognition." 2014 IEEE International Conference on Image Processing (ICIP). IEEE, 2014.
 

\bibitem{Burr2009} Burr Settles. Active learning literature survey. Computer Sciences Technical Report 1648, University of Wisconsin–Madison, 2009

 

\bibitem{Ref13} Frank Olken. Random sampling from databases. PhD thesis, University of California, Berkeley, 1993.

 \bibitem{refref24} H. Sahbi. "Lightweight Connectivity In Graph Convolutional Networks For Skeleton-Based Recognition." 2021 IEEE International Conference on Image Processing (ICIP). IEEE, 2021.
 
\bibitem{Ref14} Maria E Ramirez-Loaiza, Manali Sharma, Geet Kumar, and Mustafa Bilgic. Active learning: an empirical study of common baselines. Data mining and knowledge discovery, 31(2):287–313, 2017.



\bibitem{Ref35} Andreas Kirsch, Joost van Amersfoort, and Yarin Gal. Batchbald: Efficient and diverse batch acquisition for deep bayesian active learning, 2019.

\bibitem{refref8} H.  Sahbi.  "CNRS-TELECOM ParisTech at ImageCLEF 2013 Scalable Concept Image Annotation Task: Winning Annotations with Context Dependent SVMs." CLEF (Working Notes). 2013.

\bibitem{Ref36}  Stefan Depeweg, Jose-Miguel Hernandez-Lobato, Finale Doshi-Velez, and Steffen Udluft. Decomposition of uncertainty in bayesian deep learning for efficient and risk-sensitive learning. In International Conference on Machine Learning, pages 1184–1193. PMLR, 2018.


\bibitem{Ref40} Simon Tong and Daphne Koller. Support vector machine active learning with applications to text classification. Journal of machine learning research, 2(Nov):45–66, 2001.


\bibitem{refref1} Sahbi, H., and N.  Boujemaa. "Robust face recognition using dynamic space warping." International Workshop on Biometric Authentication. Springer, Berlin, Heidelberg, 2002.


\bibitem{Ref41}  Ashish Kapoor, Kristen Grauman, Raquel Urtasun, and Trevor Darrell.  Active learning with gaussian processes for object categorization. In 2007 IEEE 11th International Conference on Computer Vision, pages 1–8. IEEE, 2007.

\bibitem{Ref37} Sachin Ravi and Hugo Larochelle. Meta-learning for batch mode active learning. 2018. In URL https://openreview. net/forum, 2018.

 \bibitem{aaaaa5}   H.  Sahbi,  D.  Geman.  A Hierarchy of Support Vector Machines for Pattern Detection. Journal of Machine Learning Research 7 (10).


 
 
\bibitem{Ref39}  Kunkun Pang, Mingzhi Dong, Yang Wu, and Timothy Hospedales. Meta-learning transferable active learning policies by deep reinforcement learn- ing. arXiv preprint arXiv:1806.04798, 2018.

 


\bibitem{Ref47} Yi Yang, Zhigang Ma, Feiping Nie, Xiaojun Chang, and Alexander G. Hauptmann. Multi-class active learning by uncertainty sampling with diversity maximization. Int. J. Comput. Vis., 113(2):113–127, 2015.


\bibitem{refref2}S. Thiemert, H. Sahbi, and M.  Steinebach. "Using entropy for image and video authentication watermarks." Security, Steganography, and Watermarking of Multimedia Contents VIII. Vol. 6072. SPIE, 2006.



\bibitem{Ref32} Christoph Mayer and Radu Timofte. Adversarial sampling for active learning. In Proceedings of the IEEE/CVF Winter Conference on Applications of Computer Vision, pages 3071–3079, 2020.

 

\bibitem{Ref48a}  Dwarikanath Mahapatra, Behzad Bozorgtabar, Jean-Philippe Thiran, and Mauricio Reyes. Efficient active learning for image classification and seg- mentation using a sample selection and conditional generative adversarial network, 2019.

 \bibitem{refref3}H. Sahbi and N. Boujemaa. "Coarse-to-fine support vector classifiers for face detection." Object recognition supported by user interaction for service robots. Vol. 3. IEEE, 2002.
 

\bibitem{Jia2017} Jia-Jie Zhu and Jose Bento. Generative adversarial active learning. CoRR, abs/1702.07956, 2017.

\bibitem{Longlong2020}  Longlong Jing and Yingli Tian. Self-supervised visual feature learning with deep neural networks: A survey. IEEE transactions on pattern analysis and machine intelligence, 2020.

 

\bibitem{Ref50a} Yong Cheng Wu. Active learning based on diversity maximization. In Ap- plied Mechanics and Materials, volume 347, pages 2548–2552. Trans Tech Publ, 2013.

 \bibitem{aaaaa6}  L.  Wang,  H Sahbi.  Directed acyclic graph kernels for action recognition.  Proceedings of the IEEE International Conference on Computer Vision, 3168-3175.
 
 
\bibitem{Ref51a}  Sharat Agarwal, Himanshu Arora, Saket Anand, and Chetan Arora. Con- textual diversity for active learning. In European Conference on Computer Vision, pages 137–153. Springer, 2020.



\bibitem{Ref53} Yarin Gal. Uncertainty in Deep Learning. PhD thesis, University of Cambridge, 2016.

\bibitem{Ref54} Yarin Gal and Zoubin Ghahramani. Dropout as a bayesian approximation: Representing model uncertainty in deep learning, 2016.


 \bibitem{refref4} H. Sahbi. Coarse-to-fine support vector machines for hierarchical face detection. Diss. PhD thesis, Versailles University, 2003.
 
 
 
 
 
 
\bibitem{Ref56} Donggeun Yoo and In So Kweon. Learning loss for active learning. In Proceedings of the IEEE/CVF Conference on Computer Vision and Pattern Recognition, pages 93–102, 2019.


\bibitem{Ref57} Patrick Hemmer, Niklas Kuhl, and Jakob Schoffer. Deal: Deep evidential active learning for image classification. In 2020 19th IEEE International Conference on Machine Learning and Applications (ICMLA), pages 865– 870, 2020.

\bibitem{jordan2019} Jordan T Ash, Chicheng Zhang, Akshay Krishnamurthy, John Langford, and Alekh Agarwal. Deep batch active learning by diverse, uncertain gra- dient lower bounds. arXiv preprint arXiv:1906.03671, 2019.

\bibitem{refref5} N. Boujemaa,  F.  Fleuret,   V.  Gouet, and H.  Sahbi. "Visual content extraction for automatic semantic annotation of video news." In the proceedings of the SPIE Conference, San Jose, CA, vol. 6. 2004.
 

\bibitem{yoram2004}  Yoram Baram, Ran El Yaniv, and Kobi Luz. Online choice of active learning algorithms. Journal of Machine Learning Research, 5(Mar):255–291, 2004.

\bibitem{Ksenia2017}  Ksenia Konyushkova, Raphael Sznitman, and Pascal Fua. Learning active learning from data.  arXiv preprint arXiv:1703.03365, 2017.

\bibitem{ICPR2016sahbi}
H. Sahbi. "Misalignment resilient cca for interactive satellite image change detection." 2016 23rd International Conference on Pattern Recognition (ICPR). IEEE, 2016.

\bibitem{Sheng2010}  Sheng-Jun Huang, Rong Jin, and Zhi-Hua Zhou. Active learning by querying informative and representative examples. In John D. Lafferty, Christopher K. I. Williams, John Shawe-Taylor, Richard S. Zemel, and Aron Cu- lotta, editors, NIPS, pages 892–900. Curran Associates, Inc., 2010.

 \bibitem{refref6} T. Napoléon  and H.  Sahbi. "From 2D silhouettes to 3D object retrieval: contributions and benchmarking." EURASIP Journal on Image and Video Processing 2010 (2010): 1-17.
 
 
\bibitem{Naoki1998}  Naoki Abe.  Query learning strategies using boosting and bagging. Proc. of ICML98, pages 1–9, 1998.

\bibitem{reff2} Burr, Settles. "Active learning." Synthesis Lectures on Artificial Intelligence and Machine Learning 6.1 (2012).

\bibitem{refref7} X.  Li and H. Sahbi.  "Superpixel-based object class segmentation using conditional random fields." 2011 IEEE International Conference on Acoustics, Speech and Signal Processing (ICASSP). IEEE, 2011.
 
 
 


 
 
   \bibitem{refff333334} 
  Sutton, Richard S., and Andrew G. Barto. Reinforcement learning: An introduction. MIT press, 2018.
  
  \bibitem{refff333335} 
  Jin, C., Allen-Zhu, Z., Bubeck, S., \& Jordan, M. I. (2018). Is Q-learning provably efficient?. arXiv preprint arXiv:1807.03765.

 \bibitem{aaaaa3}  H.  Sahbi. Kernel PCA for similarity invariant shape recognition. Neurocomputing 70 (16-18), 3034-3045. 

\bibitem{Dosovitskiy2020}
Dosovitskiy, Alexey, et al. "An image is worth 16x16 words: Transformers for image recognition at scale." arXiv preprint arXiv:2010.11929 (2020).


\bibitem{Carl2015}
Carl Doersch, Abhinav Gupta, Alexei A. Efros. Unsupervised Visual Representation Learning by Context Prediction, arXiv:1505.05192, 2015. 

 \bibitem{refref9} M.  Jiu and H. Sahbi. "Semi supervised deep kernel design for image annotation." 2015 IEEE International Conference on Acoustics, Speech and Signal Processing (ICASSP). IEEE, 2015. 
 

\bibitem{Culotta2005}
Culotta, Aron, and Andrew McCallum. "Reducing labeling effort for structured prediction tasks." AAAI. Vol. 5. 2005.

 \bibitem{refref10}  M. Jiu  and H.  Sahbi. "Deep kernel map networks for image annotation." 2016 IEEE International Conference on Acoustics, Speech and Signal Processing (ICASSP). IEEE, 2016.
 
 


\bibitem{atwood2016diffusion} Atwood, J., Towsley,  D.: Diffusion-convolutional neural networks. In: Advances  in Neural Information Processing Systems. pp. 1993--2001 (2016)

\bibitem{bruna2013spectral} Bruna, J., Zaremba, W., Szlam, A., LeCun, Y.: Spectral networks and locally  connected networks on graphs. arXiv preprint arXiv:1312.6203  (2013)

 
 
   \bibitem{chen2017stochastic}  Chen, J., Zhu, J., Song, L.: Stochastic training of graph convolutional networks with variance reduction. arXiv preprint arXiv:1710.10568  (2017)
 
\bibitem{chen2018fastgcn} Chen, J., Ma, T., Xiao, C.: Fastgcn: fast learning with graph convolutional networks via importance sampling. arXiv preprint arXiv:1801.10247  (2018)


 \bibitem{refref12} H.  Sahbi and N.  Boujemaa. "From coarse to fine skin and face detection." Proceedings of the eighth ACM international conference on Multimedia. 2000.
 
 
\bibitem{dai2018learning} Dai, H., Kozareva, Z., Dai, B., Smola, A., Song, L.: Learning steady-states of iterative algorithms over graphs. In: International Conference on Machine Learning. pp. 1114--1122 (2018)

\bibitem{defferrard2016convolutional} Defferrard, M., Bresson, X., Vandergheynst, P.: Convolutional neural networks on graphs with fast localized spectral filtering. In: Advances in neural   information processing systems. pp. 3844--3852 (2016)
 
 
\bibitem{refref13} H.  Sahbi and F.  Fleuret.  Kernel methods and scale invariance using the triangular kernel. Diss. INRIA, 2004.


\bibitem{gao2018large} Gao, H., Wang, Z., Ji, S.: Large-scale learnable graph convolutional networks.  In: Proceedings of the 24th ACM SIGKDD International Conference on Knowledge Discovery \& Data Mining. pp. 1416--1424. ACM (2018)
 
\bibitem{gori2005new}  Gori, M., Monfardini, G., Scarselli, F.: A new model for learning in graph  domains. In: Proceedings. 2005 IEEE International Joint Conference on Neural Networks, 2005. vol.~2, pp. 729--734. IEEE (2005)

 

\bibitem{hamilton2017inductive}  Hamilton, W., Ying, Z., Leskovec, J.: Inductive representation learning on large graphs. In: Advances in Neural Information Processing Systems. pp.   1024--1034 (2017)

 \bibitem{henaff2015deep} Henaff, M., Bruna, J., LeCun, Y.: Deep convolutional networks on  graph-structured data. arXiv preprint arXiv:1506.05163  (2015)


 \bibitem{refref15} H.  Sahbi and F.  Fleuret. Scale-invariance of support vector machines based on the triangular kernel. Diss. INRIA, 2002.
 
 
\bibitem{huang2018adaptive}  Huang, W., Zhang, T., Rong, Y., Huang, J.: Adaptive sampling towards fast graph  representation learning. In: Advances in Neural Information Processing  Systems. pp. 4558--4567 (2018)

\bibitem{kipf2016semi} Kipf, T.N., Welling, M.: Semi-supervised classification with graph convolutional networks. arXiv preprint arXiv:1609.02907  (2016)


 \bibitem{aaaaa4}  H.  Sahbi,  J-Y.  Audibert,  R.  Keriven.  Context-dependent kernels for object classification. IEEE transactions on pattern analysis and machine intelligence 33 (4), 699-708.

\bibitem{levie2018cayleynets} Levie, R., Monti, F., Bresson, X., Bronstein, M.M.: Cayleynets: Graph  convolutional neural networks with complex rational spectral filters. IEEE Transactions on Signal Processing  \textbf{67}(1),  97--109 (2018)


\bibitem{li2018adaptive} Li, R., Wang, S., Zhu, F., Huang, J.: Adaptive graph convolutional neural  networks. In: Thirty-Second AAAI Conference on Artificial Intelligence (2018)

\bibitem{refref18}  L. Wang and H.  Sahbi. "Nonlinear cross-view sample enrichment for action recognition." European Conference on Computer Vision. Springer, Cham, 2014.

\bibitem{refsahbicvpr2008} H. Sahbi, P. Etyngier, J-Y. Audibert, R. Keriven. Manifold learning using robust graph laplacian for interactive image search. In CVPR 2008.

\bibitem{li2015gated} Li, Y., Tarlow, D., Brockschmidt, M., Zemel, R.: Gated graph sequence neural  networks. arXiv preprint arXiv:1511.05493  (2015)
 
\bibitem{wu2019comprehensive} Wu, Z., Pan, S., Chen, F., Long, G., Zhang, C., Yu, P.S.: A comprehensive  survey on graph neural networks. arXiv preprint arXiv:1901.00596  (2019)


\bibitem{zhang2018gaan} Zhang, J., Shi, X., Xie, J., Ma, H., King, I., Yeung, D.Y.: Gaan: Gated attention networks for learning on large and spatiotemporal graphs. arXiv  preprint arXiv:1803.07294  (2018)


 \bibitem{refref20} M. Ferecatu and H.  Sahbi. "Multi-view object matching and tracking using canonical correlation analysis." 2009 16th IEEE International Conference on Image Processing (ICIP). IEEE, 2009.

\bibitem{Ma2018} T. Ma, J. Chen, and C. Xiao, “Constrained generation of semantically valid graphs via regularizing variational autoencoders,” in Proc. of NeurIPS, 2018, pp. 7110–7121.

\bibitem{sahbi2021b} H. Sahbi. "Learning Connectivity with Graph Convolutional Networks." 2020 25th International Conference on Pattern Recognition (ICPR). IEEE, 2021.


\bibitem{Pan2018}  S. Pan, R. Hu, G. Long, J. Jiang, L. Yao, and C. Zhang, “Adversarially regularized graph autoencoder for graph embedding.” in Proc. of IJCAI 2018, pp. 2609–2615.

  
\bibitem{aaaaa1}  H. Sahbi,  L .Ballan,  G. Serra, A. Del-Bimbo. Context-dependent logo matching and recognition.  IEEE Transactions on Image Processing 22 (3), 1018-1031.


\bibitem{refffabc888}
  J. Heinonen, \textit{Lectures on Lipschitz Analysis}, Springer, 2005.

\end{thebibliography}
}
\end{document}